\pdfoutput=1

\documentclass[11pt]{article}

\usepackage[]{acl}

\usepackage{times}
\usepackage{latexsym}

\usepackage[T1]{fontenc}

\usepackage[utf8]{inputenc}

\usepackage{microtype}

\usepackage{graphicx}
\usepackage{booktabs}
\usepackage{multirow}
\usepackage{rotating}
\usepackage{amsmath}
\usepackage{amssymb}
\usepackage{bm}
\usepackage{url}
\usepackage{floatrow}
\usepackage{xcolor}
\newfloatcommand{capbtabbox}{table}[][\FBwidth]

\newcommand{\tabincell}[2]{\begin{tabular}{@{}#1@{}}#2\end{tabular}}

\title{CoMAE: A Multi-factor Hierarchical Framework for \\ Empathetic Response Generation}

\author{Chujie Zheng$^\dagger$, Yong Liu$^\ddagger$, Wei Chen$^\ddagger$, Yongcai Leng$^\ddagger$, Minlie Huang$^\dagger$\thanks{\ \ Corresponding author.} \\
  $^\dagger$The CoAI group, DCST, Institute for Artificial Intelligence, \\
  $^\dagger$State Key Lab of Intelligent Technology and Systems, \\
  $^\dagger$Beijing National Research Center for Information Science and Technology, \\
  $^\dagger$Tsinghua University, Beijing 100084, China \\
  $^\ddagger$Sogou Inc., Beijing, China \\
  {\tt chujiezhengchn@gmail.com, aihuang@tsinghua.edu.cn} \\
}

\date{}

\begin{document}
\maketitle
\begin{abstract}

  The capacity of empathy is crucial to the success of open-domain dialog systems.
  Due to its nature of multi-dimensionality, there are various factors that relate to empathy expression, such as communication mechanism, dialog act and emotion.
  However, existing methods for empathetic response generation usually either consider only one empathy factor or ignore the hierarchical relationships between different factors, leading to a weak ability of empathy modeling.
  In this paper, we propose a multi-factor hierarchical framework, CoMAE, for empathetic response generation, which models the above three key factors of empathy expression in a hierarchical way.
  We show experimentally that our CoMAE-based model can generate more empathetic responses than previous methods.
  We also highlight the importance of hierarchical modeling of different factors through both the empirical analysis on a real-life corpus and the extensive experiments.
  Our codes and used data are available at \url{https://github.com/chujiezheng/CoMAE}.

\end{abstract}

\section{Introduction}
\label{sec:introduction}

Empathy, which refers to the capacity to understand or feel what another person is experiencing \cite{rothschild2006help, read2019typology}, is a critical capability to open-domain dialog systems \cite{zhou2018design}.
As shown in previous research, empathetic conversational models can improve user satisfaction and receive more positive feedback in numerous domains \cite{klein1998computer, liu2005embedded, brave2005computers, fitzpatrick2017delivering, liu-etal-2021-towards}.
Recently, there have also been numerous works devoted to improving the dialog models' ability to understand the feelings of interlocutors \cite{rashkin-etal-2019-towards, lin-etal-2019-moel, majumder-etal-2020-mime}, which makes the dialog models more empathetic to a certain extent.

\begin{figure}[t]
  \centering
  \includegraphics[width=\linewidth]{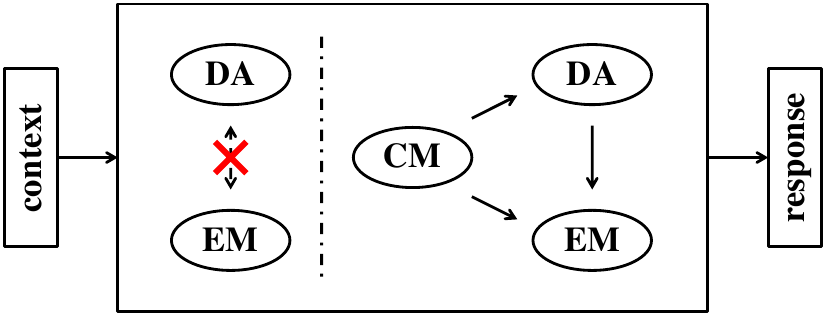}
  \caption{
    Our proposed hierarchical framework: CoMAE (right).
    The directed arrows denote dependencies.
    We also present the framework (left) of EmpTransfo \cite{zandie2020emptransfo} for comparison.
  }
  \label{fig:framework}
\end{figure}

However, empathy is a multi-dimensional construct \cite{davis1980multidimensional} rather than merely recognizing the interlocutor's emotion \cite{lin-etal-2019-moel} or emotional responding \cite{zhou2018emotional}.
It consists of two broad aspects related to \textit{cognition} and \textit{affection} \cite{omdahl2014cognitive,paiva2017empathy}.
The cognitive aspect requires understanding and interpreting the situation of the interlocutor \cite{elliott2018therapist}, which is reflected in the \textbf{dialog act} taken in the conversation \cite{de2006empathic}, such as questioning (e.g., \textit{What's wrong with it?}), consoling (e.g., \textit{You'll get through this}), etc.
The affective aspect relates to properly expressing \textbf{emotion} in reaction to the experiences and feelings shared by the interlocutor, such as admiration (e.g., \textit{Congratulations!}), sadness (e.g., \textit{I am sorry to hear that}), etc.
Very recently, \citet{sharma-etal-2020-computational} further characterizes the text-based expressed empathy based on the above two aspects as three \textbf{communication mechanisms}, which is a more higher-level and abstract factor that relates to empathy expression.

In this paper, we propose a novel framework named \textbf{CoMAE} for empathetic response generation (Section \ref{sec:comae}), which contains the aforementioned three key factors of empathy expression: \textbf{Co}mmunication \textbf{M}echanism (CM), dialog \textbf{A}ct (DA) and \textbf{E}motion (EM).
Specifically, when model these empathy factors simultaneously, we adopt a \textbf{hierarchical} way instead of following previous works that treat multiple factors independently, such like EmpTransfo \cite{zandie2020emptransfo} that considers both DA and EM (see Figure \ref{fig:framework} for comparison).
Such approaches hold the hypothesis that different factors are independent of each other, which is intuitively unreasonable.
In fact, our empirical analysis (Section \ref{sec:analysis}) on a Reddit corpus \cite{zhong-etal-2020-towards} shows that there are obvious hierarchical relationships between different factors, which confirms the soundness and necessity of hierarchical modeling.

We then devise a CoMAE-based model on top of the pre-trained language model GPT-2 \cite{radford2019language} (Section \ref{sec:methodology}), and compare the model performance with different combinations of empathy factors and hierarchical modeling.
Automatic evaluation (Section \ref{subsec:automatic}) shows that combining all the three factors hierarchically can achieve the best model performance.
Manual evaluation (Section \ref{subsec:manual}) demonstrates that our model can generate more empathetic responses than previous methods.
Extensive experiments (Section \ref{subsec:hierarchical}) further highlight the importance of hierarchical modeling in terms of the selection and realization of empathy factors.

The contributions of this paper can be summarized in three folds:
\begin{itemize}
  \item Based on the nature of multi-dimensionality of empathy expression, we propose a novel framework, CoMAE, for empathetic response generation. It hierarchically models three key factors of empathy expression: communication mechanism, dialog act and emotion.
  \item On top of GPT-2, we devise a CoMAE-based model. Experimental results show that our model can generate more empathetic responses than previous methods.
  \item We empirically analyze the necessity of hierarchical modeling, and highlight its importance especially in terms of the selection and realization of different empathy factors.
\end{itemize}

\section{Related Work}

\subsection{Factors Related to Empathy Expression}

Empathy is a complex multi-dimensional construct \cite{davis1980multidimensional} which consists of two broad aspects related to \textit{cognition} and \textit{affection} \cite{omdahl2014cognitive,paiva2017empathy}.
As shown in Section \ref{sec:introduction}, the two aspects are reflected in the dialog act (DA) taken and the emotion (EM) expressed in the conversation respectively.

Based on the theoretical definition of empathy, \citet{sharma-etal-2020-computational} characterize the text-based expressed empathy as 3 communication mechanisms (CM):
emotional reaction (ER) (e.g., \textit{I feel really sad for you}),
interpretation (IP) (e.g., \textit{This must be terrifying}, \textit{I also have similar situations}),
and exploration (EX) (e.g., \textit{Are you still feeling alone now?}).\footnote{As shown in \cite{sharma-etal-2020-computational}, the three communication mechanisms can be properly combined in one utterance. We refer the readers to their original paper for more details about the three communication mechanisms.}
These communication mechanisms are also applied in the recently proposed task of empathetic rewriting \cite{sharma2021facilitating}.

Besides, \citet{zhong-etal-2020-towards} propose that persona, which refers to the social face an individual presents to the world \cite{jung2016psychological}, has been shown to be highly correlated with personality \cite{leary2011personality}, which in turn influences empathy expression \cite{richendoller1994exploring,costa2014associations}.
While \citet{zhong-etal-2020-towards} do not explain the explicit connection between persona and empathy expression, they suggest that different speakers may have different ``styles'' for expressing empathy.

\subsection{Empathetic Response Generation}

In the past years, empathetic response generation has attracted much research interest \cite{rashkin-etal-2019-towards, lin-etal-2019-moel, majumder-etal-2020-mime, zandie2020emptransfo, sun-etal-2021-psyqa}. 
\citet{rashkin-etal-2019-towards} suggest that dialog models can generate more empathetic responses by recognizing the interlocutor's emotion.
\citet{lin-etal-2019-moel} propose to design a dedicated decoder to respond each emotion of the interlocutor, which makes the generation process more interpretable.
\citet{majumder-etal-2020-mime} adopt the idea of emotional mimicry \cite{hess2014emotional} to make the generated responses more empathetic.
Inspired by the advances in generative pre-trained language models \cite{radford2018improving, radford2019language}, EmpTransfo \cite{zandie2020emptransfo} uses GPT \cite{radford2018improving} to generate empathetic responses.

Unlike previous works that only consider the EM factor in empathy modeling, EmpTransfo takes both DA and EM into account. 
The fundamental difference of EmpTransfo from our work lies in two points: 
(1) our work further considers communication mechanism
in modeling empathy,
and (2) we analyze and explore in depth the importance of hierarchically modeling of these empathy factors.

\section{CoMAE Framework and Formulation}
\label{sec:comae}

Our proposed CoMAE framework is shown in Figure \ref{fig:framework}. 
CoMAE uses CM as a high-level factor that provides a coarse-grained guidance for empathy expression, and then takes DA and EM to achieve the fine-grained realization.
Formally, given the context $x$, CoMAE divides the generation of the empathetic response $y$ into four steps:
(1) predict \textbf{C}M $C_y$ conditioned on the context,
(2) predict D\textbf{A} $A_y$ conditioned on both the context and CoM, 
(3) predict \textbf{E}M $E_y$ based on all the conditions, 
and (4) generate the final response $y$.
The whole process is formulated as Equation \ref{equ:formulation}:
\begin{align}
    \mathbb{P} ( y, C_y, A_y, E_y | x) &= \mathbb{P} ( y | x, C_y, A_y, E_y ) \cdot \label{equ:formulation}  \\ 
    \mathbb{P} ( E_y | x, C_y &, A_y ) \mathbb{P} ( A_y | x, C_y ) \mathbb{P} ( C_y | x ). \notag
\end{align}

Note that EM is conditioned on DA, because we intuitively think the expressed emotion is the effect rather than the cause of taking some dialog act.
In the other words, one may not adopt the dialog act just for the purpose of expressing some emotion.
Hence, realizing the emotion expression as expected is also important in our task, which is the motivation of that we analyze the realization of different factors in Section \ref{subsec:hierarchical}.

It is also worth noting that while CoMAE only contains the three factors, such hierarchical framework can be naturally extended to more factors that relate to empathy expression.
For instance, \citet{zhong-etal-2020-towards} suggest that persona plays an important role in empathetic conversations.
Due to that persona may contain the information about the speaker's style of adopting DA or expressing EM, when integrating persona into empathetic response generation, being conditioned on DA and EM may lead to better performance.


\section{Data Preparation and Analysis}
\label{sec:analysis}

While no empathetic conversation corpora provide annotations of diverse empathy factors, there are abundant publicly available resources that make automatic annotation feasible.
In this section, we first introduce our used corpus and the resources and tools used in automatic annotation, then we show our empirical analysis to verify the hierarchical relationships between different empathy factors.

\subsection{Corpus}
\label{subsec:corpus}

\citet{zhong-etal-2020-towards} propose a large-scale empathetic conversation corpus\footnote{\url{https://github.com/zhongpeixiang/PEC}} 
crawled from Reddit.
It has two different domains: Happy and Offmychest.
The posts in the Happy domain mainly have positive sentiments, while those in the Offmychest domain are usually negative.
We adopted their corpus for study for two major reasons:
(1) the corpus is real-life, scalable and naturalistic rather than acted \cite{rashkin-etal-2019-towards},
and (2) the manual annotation in \cite{zhong-etal-2020-towards} shows that most of the last responses are empathetic (73\% and 61\% for Happy and Offmychest respectively).

\subsection{Annotation Resources}

\noindent \textbf{Communication Mechanism (CM)}\footnote{\url{https://github.com/behavioral-data/Empathy-Mental-Health}}
\quad
\citet{sharma-etal-2020-computational} provide two corpora annotated with CM: TalkLife (\url{talklife.co}) and Reddit (\url{reddit.com}), while only the latter is publicly accessible and we thus used the Reddit part.
Note that in their original paper, each mechanism is differentiated as three classes of ``no'', ``weak'', or ``strong''.
Due to the unbalanced distribution of three classes, we merged ``weak'' and ``strong'' into ``yes''.
Finally, we differentiated each mechanism as two classes: ``no'' or ``yes''.

\noindent \textbf{Dialog Act (DA)}\footnote{\url{https://github.com/anuradha1992/EmpatheticIntents}} 
\quad
\citet{welivita-pu-2020-taxonomy} propose a taxonomy of DA (referred as ``intent'' in the original paper) for empathetic conversations.
They first annotate 15 initial types of DA on the ED corpus \cite{rashkin-etal-2019-towards}, and finally obtain 8 high-frequency types of DA with other types merged as others (\textbf{8+others}), which are shown in Figure \ref{fig:hiera}.

\noindent \textbf{Emotion (EM)}\footnote{\url{https://github.com/google-research/google-research/tree/master/goemotions}} 
\quad
We considered the taxonomy proposed in \cite{demszky-etal-2020-goemotions}, which contains 27 emotions and a neutral one, because: 
(1) it has a wide coverage of emotion categories with clear definitions,
and (2) the annotated corpus is large-scale and also crawled from Reddit.
However, we noted that the original emotion distribution is unbalanced and the too fine-grained taxonomy may lead to the sparsity of partial emotions.
Considering the task scenario of empathetic conversation, we adopted the clustering results in \cite{demszky-etal-2020-goemotions} and modified the original taxonomy as 9 emotions and a neutral one (\textbf{9+neutral}), which are also shown in Figure \ref{fig:hiera}. 
We show the mapping between our adopted emotions and the original emotions in Appendix \ref{sec:emotion}.

\begin{table}[t]
  \centering
  \scalebox{0.8}{
    \begin{tabular}{lcccc}
        \toprule
        \textbf{Classifiers} & \textbf{Corpora} & \textbf{\# classes} & \textbf{Acc} & \textbf{F1-macro} \\
        \midrule
        \textbf{CM-ER} & Reddit & 2     & 81.2  & 76.9 \\
        \textbf{CM-IP} & Reddit & 2     & 85.7  & 85.7 \\
        \textbf{CM-EX} & Reddit & 2     & 96.4  & 92.5 \\
        \textbf{DA} & ED    & 9     & 92.0  & 87.8 \\
        \textbf{EM} & Reddit & 10    & 60.5  & 60.4 \\
        \bottomrule
    \end{tabular}%
  }
  \caption{
    Performance of the classifiers.
    ``ED'' refers to the corpus of \textsc{EmpatheticDialogues} 
    \cite{rashkin-etal-2019-towards}.
  }
  \label{tab:classifiers}
\end{table}%

\subsection{Classifiers}
\label{subsec:classifiers}

We fine-tuned the RoBERTa\footnote{\url{https://huggingface.co/roberta-base}}
\cite{liu2019roberta} classifiers for CM, DA and EM, whose performance is summarized in Table \ref{tab:classifiers}.
They all achieve reasonable performance, ensuring the quality of automatic annotation.

However, we noted that the source domain \cite{rashkin-etal-2019-towards} of the DA classifier is different from the target domain (Reddit).
To verify the quality of DA annotation, we recruited three workers from Amazon Mechanical Turk to judge whether the utterance is consistent with the annotated DA. 
From the utterances that are not annotated with ``others'', we randomly sampled 25 utterances for each DA (totally 200) to avoid the impact of unbalanced distribution.
Finally, the ratio of being judged as consistent is 0.78 with Fleiss' Kappa $\kappa=0.621$ \cite{fleiss1971measuring}, which indicates substantial agreement ($0.6 <\kappa < 0.8$) and that the automatic annotation of DA is also reliable.

\subsection{Data Filtering and Annotation}

Following the original data split of \cite{zhong-etal-2020-towards}, we first filtered those conversations where there are more than two speakers (about 15\%) to ensure that the last utterance is related to the post.
We used the aforementioned classifiers to automatically annotate each utterance with DA and EM, and annotate each final response additionally with CM.
We found that the last responses that are not annotated with any CM are more likely to be non-empathetic, thus we filtered the conversations containing such responses (about 40\%).
Finally, the sizes of Train / Valid / Test-Happy / Test-Offmychest are 125,963 / 16,371 / 11,136 / 6,413 respectively.
We show the detailed statistics of automatic annotation in Appendix \ref{sec:statistics}.

\begin{figure}[t]
  \centering
  \includegraphics[width=\linewidth]{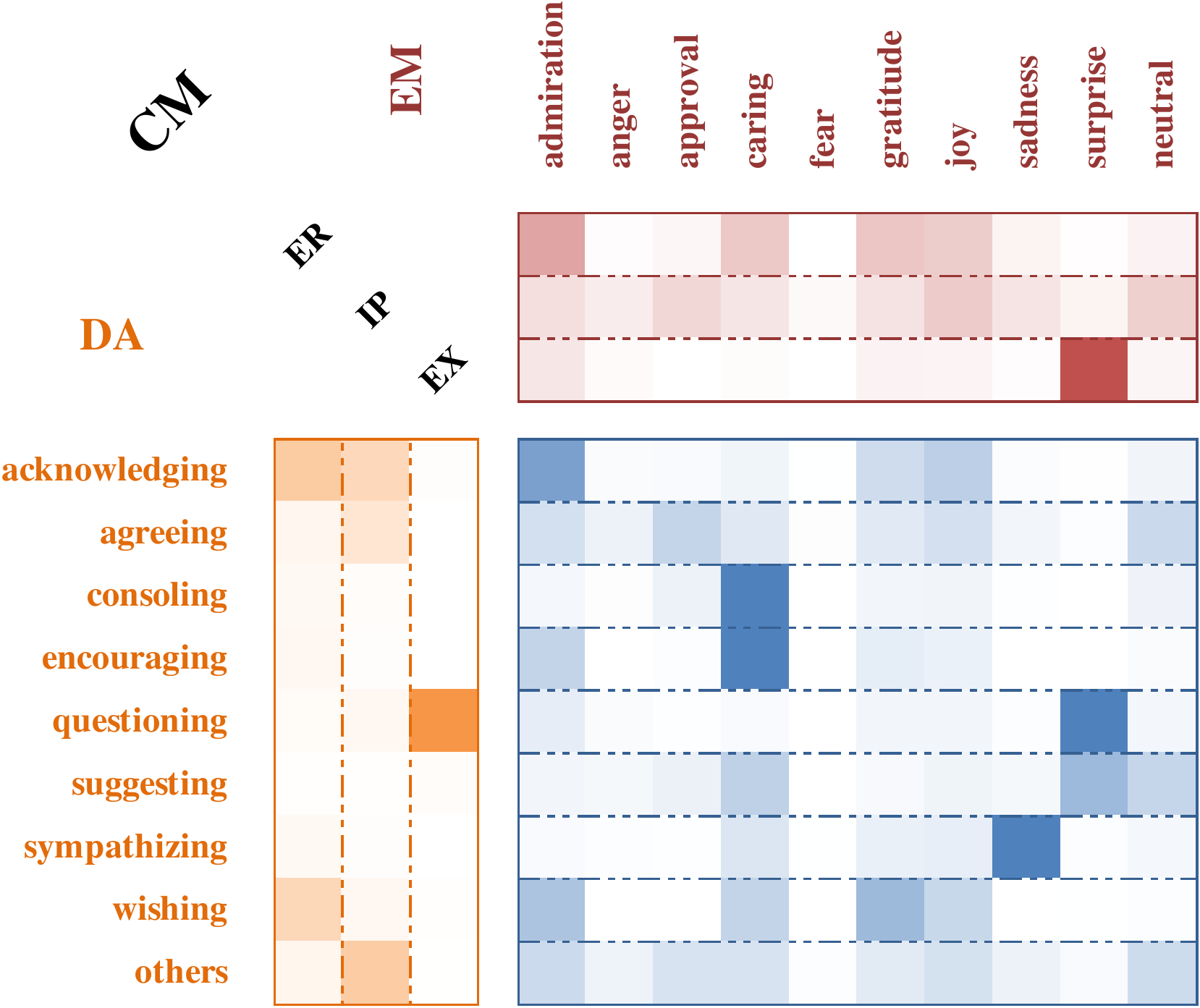}
  \caption{
    Heat maps of the conditional distributions between the three empathy factors.
    The \textcolor[RGB]{228,108,10}{\textbf{orange}} / \textcolor[RGB]{149,55,53}{\textbf{red}} / \textcolor[RGB]{79,129,189}{\textbf{blue}} maps are the distributions of \textcolor[RGB]{228,108,10}{\textbf{DA}} / \textcolor[RGB]{149,55,53}{\textbf{EM}} / \textcolor[RGB]{79,129,189}{\textbf{EM}} conditioned on \textcolor[RGB]{228,108,10}{\textbf{CM}} / \textcolor[RGB]{149,55,53}{\textbf{CM}} / \textcolor[RGB]{79,129,189}{\textbf{DA}} respectively.
  }
  \label{fig:hiera}
\end{figure}

\subsection{Analysis}
\label{subsec:analysis}

In order to verify the hierarchical relationships between the three factors, we counted the distribution frequency of each $(X,Y)$\footnote{$X$ or $Y$ is the random variable that represents CM, DA, or EM.} pair, where $(X,Y)$ is one of the three factor pairs: (CM, DA), (CM, EM), (DA, EM).
We approximated the statistical frequency of $(X,Y)$ as their joint probability distribution $\mathbb{P}(X,Y)$.
We then normalized $\mathbb{P}(X,Y)$ along the $X$ dimension to obtain the conditional distribution of $Y$ given $X$: $\mathbb{P}(Y|X)$.

Figure \ref{fig:hiera} shows the heat maps of the conditional distributions of the three factor pairs.
The heat maps reveal obvious patterns of the occurrence of $Y$ given $X$.
For instance, when one adopts the DA \textit{encouraging}, he usually expresses the EM \textit{caring} instead of \textit{approval} or \textit{joy}.
If one expresses empathy with the CM \textit{exploration (EX)}, he almost always adopts the DA \textit{questioning} and expresses the EM \textit{surprise}.
Hence, considering the hierarchical relationships between different empathy factors is reasonable and natural, and is also necessary for better empathy modeling.

\section{Methodology}
\label{sec:methodology}

\subsection{Model Architecture}
\label{subsec:architecture}

\begin{figure}[t]
  \centering
  \includegraphics[width=\linewidth]{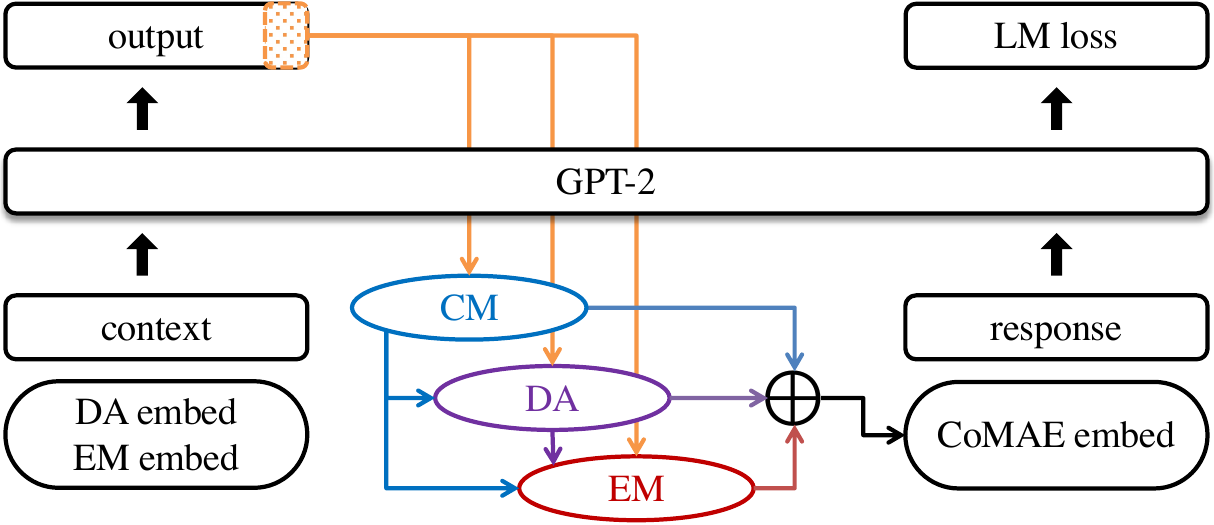}
  \caption{
    The overall architecture of our CoMAE-based model.
    The position and speaker embeddings are omitted for simplicity.
    The \textcolor[RGB]{228,108,10}{{orange dashed}} block denotes the output hidden state at the last position of the context.
  }
  \label{fig:architecture}
\end{figure}

Our devised CoMAE-based model uses GPT-2 as the backbone \cite{radford2019language}.
The overall architecture is shown in Figure \ref{fig:architecture}.

Firstly, our model takes the dialog context $x$ as input.
The context $x$ is the concatenation of history utterances: $ x = \left( u_{1}, u_{2}, \dots, u_{N} \right) $ , where $N$ is the length of dialog history.
Any two adjacent utterances are also separated by the special token \texttt{[EOS]}.
Each history utterance $u_i$ contains a sequence of tokens: $ u_{i} = \left( u_{i,1}, u_{i,2}, \dots, u_{i, l_i} \right) $, where $l_i$ is the length of $u_i$.
Each utterance $u_i$ is labeled with the corresponding speaker $k_{u_{i}} \in \{ 0, 1 \}$ (only 2 speakers).
We denote the annotated DA and EM of each utterance $u_i$ as $A_{u_{i}} \in [0, 9)$ and $E_{u_{i}} \in [0, 10)$ respectively.
Suppose that the token id and the position id of $u_{i, j}$ are denoted as $w_{u_{i, j}} \in [0, \left| \mathcal{V} \right| )$ ($\mathcal{V}$ is the vocabulary) and $p_{u_{i, j}} \in [0, 1024)$ (the maximum input length is 1024) respectively, the representation of each token $u_{i, j}$ is the summation of the following embeddings:
\begin{align}
    \bm{e}_{u_{i, j}} =\ &\bm{M}_W \left[ w_{u_{i, j}} \right] + \bm{M}_P \left[ p_{u_{i, j}} \right] +   \\ 
    & \bm{M}_K \left[ k_{u_{i}} \right] + \bm{M}_A \left[ A_{u_{i}} \right] +  \bm{M}_E \left[ E_{u_{i}} \right], \notag
\end{align}
where $\bm{M}_W \in \mathbb{R}^{ |\mathcal{V}| \times d}, \bm{M}_P \in \mathbb{R}^{ 1024 \times d}, \bm{M}_K \in \mathbb{R}^{ 2 \times d}, \bm{M}_A \in \mathbb{R}^{ 9 \times d}, \bm{M}_E \in \mathbb{R}^{ 10 \times d}$ denote the embedding matrices of word, position, speaker, DA and EM respectively, and $[\cdot]$ denotes the indexing operation.
We denote the output hidden states after feeding $x$ into the model as $\bm{H}_x \in \mathbb{R}^{ l_x \times d}$, where $l_x$ is the total length of context $x$.

Next, we use the hidden state at the last position of the context, $\bm{h}_x=\bm{H}_x \left[ -1 \right] \in \mathbb{R}^{d}$, to hierarchically predict the CM, DA and EM of the target response.
We first separately predict\footnote{In the mathematical notation used in this paper, we distinguish the ground truth value and the predicted value of a variable $X$ with the symbols $X^{*}$ and $\widehat{X}$ respectively.
} $\widehat{C}^{(i)}_y \in \left\{ 0, 1 \right\}$ for each $i \in \left\{ \mathrm{ER}, \mathrm{IP}, \mathrm{EX} \right\}$, which indicates whether to adopt the CM $i$:
\begin{align}
  \bm{h}_C^{(i)} =\ & \mathbf{F}_C^{(i)} \left(\bm{h}_x \right) \in \mathbb{R}^{d}, \label{equ:hiera_begin} \\
  \widehat{C}^{(i)}_y \sim \ & \mathbb{P} \left( \left. C^{(i)}_y \right| x \right) = \mathrm{softmax} \left( \bm{M}_C^{(i)}\bm{h}_C^{(i)} \right), \notag \\
  \widehat{C}_y =\ & \left( \widehat{C}^{(\mathrm{ER})}_y, \widehat{C}^{(\mathrm{IP})}_y, \widehat{C}^{(\mathrm{EX})}_y \right), \notag \\
  \bm{e}_{ \widehat{C}_y} =\ & \sum_{i \in \left\{ \mathrm{ER}, \mathrm{IP}, \mathrm{EX} \right\}} \bm{M}_C^{(i)} \left[ \widehat{C}^{(i)}_y \right], \label{equ:sum_strat}
\end{align}
where each $\mathbf{F}_C^{(i)}$ is a non-linear layer activated with $\mathrm{tanh}$, and each $\bm{M}_C^{(i)} \in \mathbb{R}^{ 2 \times d}$ denotes the embedding matrix of the CM $i \in \left\{ \mathrm{ER}, \mathrm{IP}, \mathrm{EX} \right\}$.
Based on the context $x$ and the predicted CMs ${ \widehat{C}_y}$, we next predict DA:
\begin{align}
  \bm{h}_A =\ &\mathbf{F}_A \left( \left[ \bm{h}_x; \bm{e}_{ \widehat{C}_y} \right] \right) \in \mathbb{R}^{d} , \label{equ:cal_da}  \\
  \widehat{A}_y \sim \ & \mathbb{P} \left( A_y \left| x, \widehat{C}_y \right. \right) =\mathrm{softmax} \left( \bm{M}_A\bm{h}_A \right), \label{equ:pred_da}
\end{align}
where $[\cdot;\cdot]$ denotes vector concatenation and $\mathbf{F}_A$ is a non-linear layer. 
Note that we share the parameters of DA embeddings with the classification head (Equation \ref{equ:pred_da}), which is consistent with the way in GPT-2 \cite{radford2019language} where the parameters of word embeddings are shared with the LM head (Equation \ref{equ:lm_head}).
EM is predicted similarly but conditioned additionally on the predicted DA $\widehat{A}_y $:
\begin{align}
  \bm{h}_E =\ &\mathbf{F}_E \left( \left[ \bm{h}_x; \bm{e}_{ \widehat{C}_y}; \bm{M}_A \left[ \widehat{A}_y \right] \right] \right) \in \mathbb{R}^{d} ,  \\
  \widehat{E}_y \sim \ & \mathbb{P} \left( E_y \left| x, \widehat{C}_y, \widehat{A}_y \right. \right) = \mathrm{softmax} \left( \bm{M}_E\bm{h}_E \right),  \label{equ:pred_emo}
\end{align}
where $\mathbf{F}_E$ is also a non-linear layer.

Finally, we add all the factors to obtain the fused embedding $\bm{e}_\text{CoMAE}$ that controls the empathy expression of the response:
\begin{align}
    \bm{e}_\text{CoMAE} = \bm{e}_{ \widehat{C}_y} + \bm{M}_A \left[ \widehat{A}_y \right] + \bm{M}_E \left[ \widehat{E}_y \right] .  \notag 
\end{align}
The embedding of each input token $\widehat{y}_t$ in the response is as follows:
\begin{align}
    \bm{e}_{ \widehat{y}_{t}} =\ &\bm{M}_W \left[ w_{\widehat{y}_t} \right] + \bm{M}_P \left[ p_{\widehat{y}_t} \right] +  \\ 
    & \bm{M}_K \left[ k_y \right] + \bm{e}_\text{CoMAE}. \notag
\end{align}
Suppose that the output hidden state corresponding to $\widehat{y}_t$ is $\bm{s}_{t}$, then we predict the next token $\widehat{y}_{t+1}$ through the LM head:
\begin{align}
    \widehat{y}_{t+1} &\sim \mathbb{P} \left( y_{t+1} \left| \widehat{y}_{\le t}; x, \widehat{C}_y, \widehat{A}_y, \widehat{E}_y \right. \right) \label{equ:lm_head} \\
    &= \mathrm{softmax} \left( \bm{M}_W \bm{s}_t \right), \notag
\end{align}
where the parameters of the LM head are shared with the word embedding matrix $\bm{M}_W$.

\subsection{Training}
\label{subsec:training}

The optimization object contains two parts.
One part is the negative log likelihood loss $\mathcal{L}_\mathrm{NLL}$ of the target response:
\begin{align}
    \mathcal{L}_\mathrm{NLL} = - \frac{1}{l_y} \sum_{t=1}^{l_y} \ln \mathbb{P} \left( y^{*}_{t} \left| y^{*}_{< t}; x, C^{*}_y, A^{*}_y, E^{*}_y \right. \right), \notag 
\end{align}
where $l_y$ is the length of the golden response.
The other part is the prediction losses of CM $\mathcal{L}_C$, DA $\mathcal{L}_A$, and EM $\mathcal{L}_E$:
\begin{align}
    \mathcal{L}_C = \ & - \sum_{i \in \left\{ \mathrm{ER}, \mathrm{IP}, \mathrm{EX} \right\}} \ln \mathbb{P} \left( \left. C^{(i)*}_y \right| x \right), \label{equ:loss_c} \\
    \mathcal{L}_A = \ & - \ln \mathbb{P} \left( A^{*}_y \left| x, {C}^{*}_y \right. \right), \label{equ:loss_a} \\
    \mathcal{L}_E = \ & - \ln \mathbb{P} \left( E^{*}_y \left| x, {C}^{*}_y, {A}^{*}_y \right. \right). \label{equ:loss_e}
\end{align}
The complete optimization object is the summation of the above losses:
$
    \mathcal{L} = \mathcal{L}_\mathrm{NLL} + \lambda \left( \mathcal{L}_C  + \mathcal{L}_A + \mathcal{L}_E \right) 
$, 
where $\lambda$ is the weight of the prediction losses.
We set $\lambda$ to 1.0 in our experiments.

\subsection{Discussion}

It is worth noting that the supervision signals of predictions (from Equation \ref{equ:loss_c} to \ref{equ:loss_e}) combined with hierarchical modeling (from Equation \ref{equ:hiera_begin} to \ref{equ:pred_emo}) enable the model to establish the connections between the embeddings of the three factors. 
For instance, in Equation \ref{equ:pred_da}, the embedding matrix of DA, $\bm{M}_A$, is multiplied with $\bm{h}_A$, which explicitly contains the information of the embedding matrices of CM, $\bm{M}_C^{(i)}$ (Equation \ref{equ:sum_strat} and \ref{equ:cal_da}).
The case of Equation \ref{equ:pred_emo} is similar, where $\bm{M}_E$ is multiplied with $\bm{h}_E$ that directly relates to $\bm{M}_C^{(i)}$ and $\bm{M}_A$.

Hence, consider two models where one uses hierarchical modeling and the other does not (predicting each factor separately).
When the two models are fed with the same empathy factors, saying the triplet $(C_y, A_y, E_y)$ is designated validly, we can expect that the former model has better performance than the latter one.
This conjecture will be verified in the automatic evaluation (Section \ref{subsec:automatic}).

\section{Experiments}

\subsection{Compared Models}

We investigated the model performance with different combinations of empathy factors and hierarchical modeling:

\noindent (1) \textbf{Vanilla}: the GPT-2 model directly fine-tuned on the corpus without adding any empathy factor;

\noindent (2) \textbf{+CM}, \textbf{+DA}, \textbf{+EM}: the GPT-2 models equipped with one of the three factors;

\noindent (3) \textbf{CM} $\bm{||}$ \textbf{DA}, \textbf{CM} $\bm{||}$ \textbf{EM}, \textbf{DA} $\bm{||}$ \textbf{EM}, \textbf{CM} $\bm{||}$ \textbf{DA} $\bm{||}$ \textbf{EM}: the models equipped with two or all of the three factors, but predicting each factor separately without hierarchical modeling;

\noindent (4) \textbf{CM} $\bm{\to}$ \textbf{DA}, \textbf{CM} $\bm{\to}$ \textbf{EM}, \textbf{DA} $\bm{\to}$ \textbf{EM}, \textbf{CM} $\bm{\to}$ \textbf{DA} $\bm{\to}$ \textbf{EM}: the models that are similar to (3) but utilize the hierarchical relationships, where $\to$ denotes dependency.

Note that the baseline {DA} ${||}$ {EM} is consistent with EmpTransfo\footnote{
{DA} ${||}$ {EM} has the same input representation except the speaker embeddings as EmpTransfo, but is instead fine-tuned from GPT-2 rather than GPT. 
Besides, we did not adopt the next sentence prediction (NSP) task as in \cite{zandie2020emptransfo}, because we empirically found that adding NSP leads to worse performance.
} \cite{zandie2020emptransfo}, and {CM} ${\to}$ {DA} ${\to}$ {EM} is exactly our devised model described in Section \ref{subsec:architecture}.

\subsection{Implementation Details}

All the models were implemented with PyTorch\footnote{\url{https://pytorch.org/}} \cite{paszke2019pytorch} and the Transformers library\footnote{\url{https://github.com/huggingface/transformers}} \cite{wolf-etal-2020-transformers}.
We used the pre-trained GPT-2 with the size of 117M parameters (768 hidden sizes, 12 heads, 12 layers) for all the models.
The responses were decoded by Top-$p$ sampling with $p=0.9$ and the temperature $\tau=0.7$ \cite{holtzman2019curious}. 
We trained all the models with Adam \cite{kingma2014adam} optimizer with $\beta_1 = 0.9$ and $\beta_2 = 0.999$.
The learning rate was $10^{-4}$ and was dynamically changed using the linear warmup \cite{popel2018training} with 4000 warmup steps.
All the models were fine-tuned for 5 epochs with the batch size 16 on one NVIDIA RTX 2080Ti GPU.
We selected the checkpoint for each model where the model obtains the lowest perplexity score on the Valid set. 

\subsection{Automatic Evaluation}
\label{subsec:automatic}

The automatic evaluation uses the golden responses as reference to evaluate the responses generated by models.
However, when the responses are generated based on the predicted CM / DA / EM, it is not appropriate to compare the generated responses with the reference ones \cite{liu-etal-2016-evaluate}.
Thus, in automatic evaluation we only considered the setting where the models are fed with the ground truth empathy factors.
The results where the generated responses are based on the predicted factors will be analyzed in the later experiments.

The automatic metrics we adopted include perplexity (\textbf{PPL}), BLEU-2 (\textbf{B-2}) \cite{papineni-etal-2002-bleu}, ROUGE-L (\textbf{R-L}) \cite{lin-2004-rouge}, 
and the BOW Embedding-based \cite{liu-etal-2016-evaluate} \textbf{Greedy} matching score.
The metrics except PPL were calculated with an NLG evaluation toolkit\footnote{\url{https://github.com/Maluuba/nlg-eval}} \cite{sharma2017relevance}, where the generated responses were tokenized with NLTK\footnote{\url{https://www.nltk.org/}} \cite{loper2002nltk}.

\begin{table}[t]
  \centering
  \scalebox{0.75}{
    \begin{tabular}{clcccc}
        \toprule
              & \textbf{Models} & \textbf{PPL} & \textbf{B-2} & \textbf{R-L} & \textbf{Greedy} \\
        \midrule
        \multirow{12}[6]{*}{\begin{sideways}\textbf{Happy}\end{sideways}} & \textbf{Vanilla} & 18.82 & 5.95* & 15.00* & 66.09* \\
              & \textbf{+CM} & 18.21 & 6.67* & 17.64* & 66.95* \\
              & \textbf{+DA} & 18.01 & 7.18* & 18.09* & 67.35* \\
              & \textbf{+EM} & 17.88 & 7.51* & 18.27* & 67.78* \\
    \cmidrule{2-6}          & \textbf{CM} $\bm{||}$ \textbf{DA}  & 17.83 & 7.76* & 18.85* & 67.78* \\
              & \textbf{CM} $\bm{||}$ \textbf{EM}  & 17.57 & 8.17* & 19.58* & 68.25* \\
              & \textbf{DA} $\bm{||}$ \textbf{EM}  & 17.38 & 8.37* & 19.91* & 68.59* \\
              & \textbf{CM} $\bm{||}$ \textbf{DA} $\bm{||}$ \textbf{EM}  & 17.26 & 9.21  & 20.75 & 68.86 \\
    \cmidrule{2-6}          & \textbf{CM} $\bm{\to}$ \textbf{DA}  & 17.69 & 7.95* & 18.96* & 67.79* \\
              & \textbf{CM} $\bm{\to}$ \textbf{EM}  & 17.45 & 8.04* & 19.49* & 68.08* \\
              & \textbf{DA} $\bm{\to}$ \textbf{EM}  & 17.28 & 8.73* & 20.09* & 68.59* \\
              & \textbf{CM} $\bm{\to}$ \textbf{DA} $\bm{\to}$ \textbf{EM}  & \textbf{17.02} & \textbf{9.44} & \textbf{20.76} & \textbf{68.92} \\
        \midrule
        \multirow{12}[6]{*}{\begin{sideways}\textbf{Offmychest}\end{sideways}} & \textbf{Vanilla} & 22.11 & 5.66* & 13.75* & 68.40* \\
              & \textbf{+CM} & 21.44 & 6.65* & 17.62* & 69.68* \\
              & \textbf{+DA} & 21.34 & 7.11* & 17.44* & 69.67* \\
              & \textbf{+EM} & 21.26 & 6.75* & 17.40* & 69.63* \\
    \cmidrule{2-6}          & \textbf{CM} $\bm{||}$ \textbf{DA}  & 21.07 & 7.56* & 18.41* & 70.16* \\
              & \textbf{CM} $\bm{||}$ \textbf{EM}  & 20.83 & 7.78* & 18.97* & 70.34* \\
              & \textbf{DA} $\bm{||}$ \textbf{EM}  & 20.85 & 7.48* & 18.49* & 70.19* \\
              & \textbf{CM} $\bm{||}$ \textbf{DA} $\bm{||}$ \textbf{EM}  & 20.63 & 8.23  & 19.32 & 70.54 \\
    \cmidrule{2-6}          & \textbf{CM} $\bm{\to}$ \textbf{DA}  & 20.87 & 7.70* & 18.58* & 70.33* \\
              & \textbf{CM} $\bm{\to}$ \textbf{EM}  & 20.72 & 7.71* & 18.63* & 70.31* \\
              & \textbf{DA} $\bm{\to}$ \textbf{EM}  & 20.68 & 7.89* & 18.66* & 70.25* \\
              & \textbf{CM} $\bm{\to}$ \textbf{DA} $\bm{\to}$ \textbf{EM}  & \textbf{20.35} & \textbf{8.35} & \textbf{19.54} & \textbf{70.68} \\
        \bottomrule
    \end{tabular}%
  }
  \caption{
    Results of automatic evaluation.
    The best results are in \textbf{bold}.
    {DA} ${||}$ {EM} is consistent with EmpTransfo \cite{zandie2020emptransfo}. 
    {CM} ${\to}$ {DA} ${\to}$ {EM} is our devised model described in Section \ref{subsec:architecture}.
    Scores that are significantly worse than the best scores are marked with * (Student's t-test, $p$-value $<0.05$).
  }
  \label{tab:automatic}%
\end{table}%

Results are shown in Table \ref{tab:automatic}.
We analyze the results from the following three perspectives:

\noindent \textbf{General Performance} \quad
Our model achieves the best performance on all the metrics on both domains, and most of the advantages over the competitors are statistically significant.

\noindent \textbf{Impact of Empathy Factors} \quad
The model performance vary from different combinations of empathy factors.
First, considering more empathy factors always leads to better performance (e.g., {CM} ${\to}$ {DA} ${\to}$ {EM} $>$ {CM} ${\to}$ {EM} $>$ {+EM} $>$ Vanilla).
Second, EM brings the most gains to the model performance among the three factors.
It may be because emotion is the most explicit factor that influences empathy expression 
\cite{sharma-etal-2020-computational}.
In contrast, CM brings fewer gains than DA and EM.
The reason may be that CM provides a high-level but coarse-grained guidance for empathetic response generation, lacking a fine-grained control like DA or EM.
While the responses in the corpus of \cite{zhong-etal-2020-towards} are not too long ($\le$ 30 words), we believe that CM plays an important role in generating longer empathetic responses, which may require the planning of multiple methanisms and more diverse usage of DA and EM.

\noindent \textbf{Impact of Hierarchical modeling} \quad
We noticed that for almost all the models that adopt multiple empathy factors, hierarchical modeling always leads to better performanc (e.g., {CM} ${\to}$ {DA} ${\to}$ {EM} $>$ {CM} ${||}$ {DA} ${||}$ {EM}, 
{DA} ${\to}$ {EM} $>$ {DA} ${||}$ {EM}).
This phenomenon is not trivial because the models with or without hierarchical modeling are all fed with the same empathy factors as the reference responses.
It confirms our conjecture in Section \ref{subsec:training} that hierarchical modeling can establish the connections between the embeddings of different factors, thus leading to a better capacity of empathy modeling.
However, (CM, EM) is an exception.
It may be due to that the pair (CM, EM) has a weaker correlation (the lowest manual information, Section \ref{subsec:analysis}) than other pairs.

\subsection{Manual Evaluation}
\label{subsec:manual}

In manual evaluation, the models generate responses based on the empathy factors sampled from the predicted probability distributions.
When sampling DA or EM, we used the Top-$p$ filtering with $p=0.9$ \cite{holtzman2019curious} to ensure the validness of the sampled results.

The manual evaluation is based on pair-wise comparison, and the metrics for manual evaluation include: 
\textbf{Fluency} (which response has better fluency and readability),
\textbf{Coherence} (which response has better coherence and higher relevance to the context),
and \textbf{Empathy} (which response shows better understanding of the partner's experiences and feelings, and which response expresses empathy in the way that the annotators prefer).
The pair-wise comparison is conducted between three pairs of models: 
(1) {CM} ${\to}$ {DA} ${\to}$ {EM} vs. {DA} ${\to}$ {EM}, 
(2) {CM} ${\to}$ {DA} ${\to}$ {EM} vs. {CM} ${||}$ {DA} ${||}$ {EM}, 
and (3) {DA} ${\to}$ {EM} vs. {DA} ${||}$ {EM}.
We randomly sampled 100 conversations from each test set of two domains (totally 200), and recruited three workers from Amazon Mechanical Turk for annotation.

\begin{table}[t]
  \centering
  \scalebox{0.8}{
    \begin{tabular}{ccccc}
        \toprule
        \textbf{Comparisons} & \textbf{Metrics} & \textbf{Win} & \textbf{Lose} & $\bm{\kappa}$ \\
        \midrule
        \multirow{3}{*}{\tabincell{c}{\textbf{CM} $\bm{\to}$ \textbf{DA} $\bm{\to}$ \textbf{EM}  \\ \textbf{vs.} \\ \textbf{DA} $\bm{\to}$ \textbf{EM}}} & Flu   & 33.3  & {34.8}  & 0.330 \\
              & Coh   & 35.3  & {39.3}  & 0.431 \\
              & Emp*  & \textbf{39.3}  & 32.3  & 0.402 \\
        \midrule
        \multirow{3}{*}{\tabincell{c}{\textbf{CM} $\bm{\to}$ \textbf{DA} $\bm{\to}$ \textbf{EM}  \\ \textbf{vs.} \\ \textbf{CM} $\bm{||}$ \textbf{DA} $\bm{||}$ \textbf{EM}}} & Flu   & {37.3}  & 34.5  & 0.383 \\
              & Coh*  & \textbf{41.6}  & 33.4  & 0.412 \\
              & Emp   & {43.4}  & 39.6  & 0.416 \\
        \midrule
        \multirow{3}{*}{\tabincell{c}{\textbf{DA} $\bm{\to}$ \textbf{EM}  \\ \textbf{vs.} \\ \textbf{DA} $\bm{||}$ \textbf{EM}}} & Flu   & 36.2  & {38.5}  & 0.381 \\
              & Coh   & \textbf{40.0}  & 35.7  & 0.523 \\
              & Emp   & {44.7}  & 42.0  & 0.497 \\
        \bottomrule
    \end{tabular}%
  }
  \caption{
    Results of manual evaluation.
    Ties are not shown. 
    The metrics with significant gaps are marked with * (sign test, $p$-value $<0.05$).
    $\kappa$ denotes Fleiss’ Kappa, whose values indicate fair agreement ($0.2 < \kappa < 0.4$) or moderate agreement ($0.4 < \kappa < 0.6$).
  }
  \label{tab:manual}%
\end{table}%

\begin{table}[t]
  \centering
  \scalebox{0.8}{
    \begin{tabular}{clcccc}
        \toprule
              & $\bm{X, Y}$ & \boldmath{}\textbf{Acc of $X$}\unboldmath{} & \textbf{Prop.} & \multicolumn{2}{c}{\boldmath{}\textbf{Hits@1/3 of $Y$}\unboldmath{}} \\
        \midrule
        \multirow{6}[6]{*}{\begin{sideways}\textbf{Happy}\end{sideways}} & \textbf{CM} $\bm{||}$ \textbf{DA}  & 69.5  & \multirow{2}[1]{*}{68.9} & 46.1* & 81.5* \\
              & \textbf{CM} $\bm{\to}$ \textbf{DA}  & 70.2  &       & \textbf{49.5} & \textbf{85.1} \\
    \cmidrule{2-6}          & \textbf{CM} $\bm{||}$ \textbf{EM}  & 69.5  & \multirow{2}[1]{*}{68.9} & 42.3  & 80.1* \\
              & \textbf{CM} $\bm{\to}$ \textbf{EM}  & 70.4  &       & \textbf{42.8} & \textbf{82.7} \\
    \cmidrule{2-6}          & \textbf{DA} $\bm{||}$ \textbf{EM}  & 40.1  & \multirow{2}[1]{*}{34.6} & 50.3* & 86.5* \\
              & \textbf{DA} $\bm{\to}$ \textbf{EM}  & 40.0  &       & \textbf{53.5} & \textbf{89.7} \\
        \midrule
        \multirow{6}[6]{*}{\begin{sideways}\textbf{Offmychest}\end{sideways}} & \textbf{CM} $\bm{||}$ \textbf{DA}  & 48.4  & \multirow{2}[1]{*}{45.2} & 41.3* & 67.9* \\
              & \textbf{CM} $\bm{\to}$ \textbf{DA}  & 49.2  &       & \textbf{45.9} & \textbf{75.1} \\
    \cmidrule{2-6}          & \textbf{CM} $\bm{||}$ \textbf{EM}  & 45.7  & \multirow{2}[1]{*}{42.9} & 47.2* & 74.2* \\
              & \textbf{CM} $\bm{\to}$ \textbf{EM}  & 46.1  &       & \textbf{50.3} & \textbf{77.2} \\
    \cmidrule{2-6}          & \textbf{DA} $\bm{||}$ \textbf{EM}  & 35.0  & \multirow{2}[1]{*}{30.7} & 60.5* & 84.8* \\
              & \textbf{DA} $\bm{\to}$ \textbf{EM}  & 34.9  &       & \textbf{70.2} & \textbf{88.3} \\
        \bottomrule
    \end{tabular}%
  }
  \caption{
    Results of the Hits@1/3 of predicting $Y$ given that $X$ is predicted rightly.
    ``Prop.'' denotes the proportion of the cases where both models ${X \ || \ Y}$ and ${X \to Y}$ predict $X$ rightly.
    Scores that are significantly improved after using hierarchical modeling are marked with * (sign test, $p$-value $<0.001$).
  }
  \label{tab:prediction}%
\end{table}%

Results are shown in Table \ref{tab:manual}.
From all the three pairs, we find that the responses generated by these GPT-2-based models have similar fluency.
The results of (1) indicate that further considering CM can significantly improve the empathy of generated responses, while the coherence may slightly decrease.
It may be because that the communication mechanisms like interpretation sometimes lead to the responses that are less relevant to the contexts (especially those sharing experiences).
The results of (2) and (3) indicate that hierarchical modeling improves the coherence of generated responses. 
The more empathy factors are modeled, the larger improvement can be obtained.

\subsection{Further Analysis of Hierarchical modeling}
\label{subsec:hierarchical}

To give further insights of the superiority of hierarchical modeling, we analyzed (1) the prediction and (2) the realization of empathy factors.

\noindent \textbf{Prediction}\quad
For each pair $(X,Y)$ in (CM, DA), (CM, EM), (DA, EM), we paired the models ${X \ || \ Y}$ and ${X \to Y}$ for comparison.
Our purpose is to observe whether the prediction of $X$ improves that of $Y$ after using hierarchical modeling.
Note that when taking the ground truth as reference, it is not appropriate to directly judge the prediction accuracy by comparing $\widehat{Y}$ and $Y^{*}$ if $\widehat{X} \neq X^{*}$.
We thus computed the conditional probability that $Y$ is predicted rightly given that $X$ is predicted rightly: $\mathbb{P} \left( \left. \widehat{Y} = Y^{*} \right| \widehat{X} = X^{*} \right)$.

\begin{table}[t]
  \centering
  \scalebox{0.8}{
    \begin{tabular}{clccc}
        \toprule
              & \textbf{Models} & \textbf{CM} & \textbf{DA} & \textbf{EM} \\
        \midrule
        \multirow{8}[8]{*}{\begin{sideways}\textbf{Happy}\end{sideways}} & \textbf{CM} $\bm{||}$ \textbf{DA}  & 69.6* & 76.2* & - \\
              & \textbf{CM} $\bm{\to}$ \textbf{DA}  & \textbf{79.3} & \textbf{83.6} & - \\
    \cmidrule{2-5}          & \textbf{CM} $\bm{||}$ \textbf{EM}  & 73.8* & -     & 78.0* \\
              & \textbf{CM} $\bm{\to}$ \textbf{EM}  & \textbf{76.6} & -     & \textbf{82.4} \\
    \cmidrule{2-5}          & \textbf{DA} $\bm{||}$ \textbf{EM}  & -     & 77.5* & 75.0* \\
              & \textbf{DA} $\bm{\to}$ \textbf{EM}  & -     & \textbf{87.3} & \textbf{85.7} \\
    \cmidrule{2-5}          & \textbf{CM} $\bm{||}$ \textbf{DA} $\bm{||}$ \textbf{EM}  & 68.5* & 70.3* & 71.9* \\
              & \textbf{CM} $\bm{\to}$ \textbf{DA} $\bm{\to}$ \textbf{EM}  & \textbf{76.7} & \textbf{83.7} & \textbf{81.2} \\
        \midrule
        \multirow{8}[8]{*}{\begin{sideways}\textbf{Offmychest}\end{sideways}} & \textbf{CM} $\bm{||}$ \textbf{DA}  & 61.8* & 65.6* & - \\
              & \textbf{CM} $\bm{\to}$ \textbf{DA}  & \textbf{71.4} & \textbf{74.8} & - \\
    \cmidrule{2-5}          & \textbf{CM} $\bm{||}$ \textbf{EM}  & 65.4* & -     & 66.1* \\
              & \textbf{CM} $\bm{\to}$ \textbf{EM}  & \textbf{71.1} & -     & \textbf{74.6} \\
    \cmidrule{2-5}          & \textbf{DA} $\bm{||}$ \textbf{EM}  & -     & 63.7* & 58.3* \\
              & \textbf{DA} $\bm{\to}$ \textbf{EM}  & -     & \textbf{79.5} & \textbf{75.1} \\
    \cmidrule{2-5}          & \textbf{CM} $\bm{||}$ \textbf{DA} $\bm{||}$ \textbf{EM}  & 59.0* & 60.8* & 58.9* \\
              & \textbf{CM} $\bm{\to}$ \textbf{DA} $\bm{\to}$ \textbf{EM}  & \textbf{70.7} & \textbf{76.2} & \textbf{72.6} \\
        \bottomrule
    \end{tabular}%
  }
  \caption{
    Realization scores.
    All the scores are significantly improved after using hierarchical modeling (sign test, $p$-value $<0.00001$).
  }
  \label{tab:control}%
\end{table}%

Results are shown in Table \ref{tab:prediction}.
While the accuracy of predicting $X$ of ${X \ || \ Y}$ and ${X \to Y}$ is close, the prediction of $Y$ is significantly enhanced by hierarchical modeling.
The results demonstrate that hierarchical modeling enables the model to select more proper empathy factors.

\noindent \textbf{Realization}\quad
Recall that in manual evaluation, the models generate a response based on the sampled empathy factors $\widehat{C}_y, \widehat{A}_y, \widehat{E}_y$.
To verify whether these factors are well realized, we used the classifiers in Section \ref{subsec:classifiers} to identify the empathy factors displayed in the generated responses.
Suppose that the identification results are $\widetilde{Z}_y, \forall Z \in \{ C, A, E \}$, we computed the ratio of $\widehat{Z}_y=\widetilde{Z}_y$ as the realization score of $Z$.

Results are shown in Table \ref{tab:control}.
The realization of all the factors is significantly improved by hierarchical modeling.
It is intuitive because hierarchical modeling can avoid the cases where the sampled factors are inappropriate or even conflicting, thus reducing the noise of empathy factors in response generation.

\subsection{Case Study}

We show the generated responses with different empathy factors in Figure \ref{fig:case}.
The adoption of the CM \textit{emotional reaction} causes our model to express the same EM \textit{admiration} (\textit{i'm proud of you!}) as DA ${\to}$ EM (\textit{good for you, man!}), while the two models generate the same sentence (\textit{keep it up!}) when taking the DA \textit{encouraging}.
However, the further adoption of the CM \textit{interpretation} causes our model to further share its own experiences and feelings (\textit{i have been sober for about 10 years, and it's the best feeling ever}).
As a result, with the enhancement of multiple empathy factors, the response generated by our model is more engaging and empathetic while maintaining the coherence.

Besides, we noticed another phenomenon occurring when all the three CMs are adopted.
In this case, the three CMs are usually represented separately in different sentences (e.g., \textit{I am so happy for you! I also had tried to be sober but failed. How did you make it?}), which is consistent with the results of empathetic rewriting \cite{sharma2021facilitating}.
Recall that we add the same CoMAE embeddings for all the tokens in the response during generation (Section \ref{subsec:architecture}).
Such uniform operation seems non-optimal for the nonuniform realization of different CMs, especially when generating a longer empathetic response that contains multiple sentences with different CMs, DAs or EMs. 
We believe there is still much room of improvement when applying our CoMAE framework to longer response generation, like combining CoMAE's multi-factor hierarchical modeling with planning-based dialog generation methods \cite{ghazarian2021discol}.

\begin{figure}[t]
  \centering
  \scalebox{0.8}{
    \begin{tabular}{p{23em}}
        \toprule
        \textbf{Post} \\ you might remember me posting here when i had less than a month sober a little while back. well, yesterday i hit 100 days without alcohol and celebrated by solo hiking my state's tallest mountain! \\
        \toprule
        \textbf{Golden} \\ ok that is an awesome pic! love it and the story thank you! \\
        \midrule
        \textbf{EM} \\ you're doing great! \\
        \midrule
        \textbf{DA $\bm{\to}$ EM} \\ good for you, man! keep it up! \\
        \midrule
        \textbf{CM $\bm{\to}$ DA $\bm{\to}$ EM (Ours)} \\ \textit{i'm proud of you! i have been sober for about 10 years, and it's the best feeling ever. keep it up!} \\
        \bottomrule
    \end{tabular}%
  }
  \caption{
    Responses generated with different empathy factors.
    All the generated responses express the EM \textit{admiration}.
    \textbf{DA $\bm{\to}$ EM} takes the DA \textit{encouraging}.
    \textbf{Ours} further adopts the CM \textit{emotional reaction} and \textit{interpretation}.
  }
  \label{fig:case}%
\end{figure}%

\section{Conclusion}

In this paper, we present a multi-factor hierarchical framework CoMAE for empathetic response generation.
It contains three key factors of empathy expression: communication mechanism, dialog act and emotion, and models these factors in a hierarchical way.
With our devised CoMAE-based model, we empirically demonstrate the effectiveness of these empathy factors, as well as the necessity and importance of hierarchical modeling.

As future work, the CoMAE framework can be naturally extended to more factors that relate to empathy expression, such as persona \cite{zhong-etal-2020-towards}, by exploring the hierarchical relationships between different factors.


\section*{Acknowledgments}

This work was supported by the NSFC projects (Key project with No. 61936010 and regular project with No. 61876096). 
This work was also supported by the Guoqiang Institute of Tsinghua University, with Grant No. 2019GQG1 and 2020GQG0005.

\bibliography{anthology,acl}
\bibliographystyle{acl_natbib}

\newpage
\appendix

\section{Emotion Mapping}
\label{sec:emotion}

In the original paper of \cite{demszky-etal-2020-goemotions}\footnote{\url{https://arxiv.org/abs/2005.00547v2}}, the authors provide the hierarchical clustering results of the 27 emotions (Figure 2 in their paper), which reflect the nested structure of their proposed emotion taxonomy.
Based on the clustering results, we merged the emotions that are highly correlated with each other, and the mapping between our adopted emotions and the original emotions is shown in Table \ref{tab:mapping}.

\begin{table}[h]
  \centering
  \scalebox{0.8}{
    \begin{tabular}{p{5em}p{15em}}
        \toprule
        \textbf{Ours} & \textbf{Original} \\
        \midrule
        {admiration} & {admiration, pride} \\
        \cmidrule{1-2}
        {anger} & {anger, annoyance, disgust, disapproval} \\
        \cmidrule{1-2}
        {approval} & {approval, realization} \\
        \cmidrule{1-2}
        {caring} & {caring, desire, optimism} \\
        \cmidrule{1-2}
        {fear} & {fear, nervousness} \\
        \cmidrule{1-2}
        {gratitude} & {gratitude, relief} \\
        \cmidrule{1-2}
        {joy} & {joy, amusement, excitement, love} \\
        \cmidrule{1-2}
        {sadness} & {sadness, disappointment, embarrassment, grief, remorse} \\
        \cmidrule{1-2}
        {surprise} & {surprise, confusion, curiosity} \\
        \bottomrule
    \end{tabular}%
  }
  \caption{
    Mapping between our adopted emotions and the original emotions in \cite{demszky-etal-2020-goemotions}.
  }
  \label{tab:mapping}%
\end{table}%

\section{Statistics of Annotation}
\label{sec:statistics}

\begin{figure}[h]
  \centering
  \includegraphics[width=\linewidth]{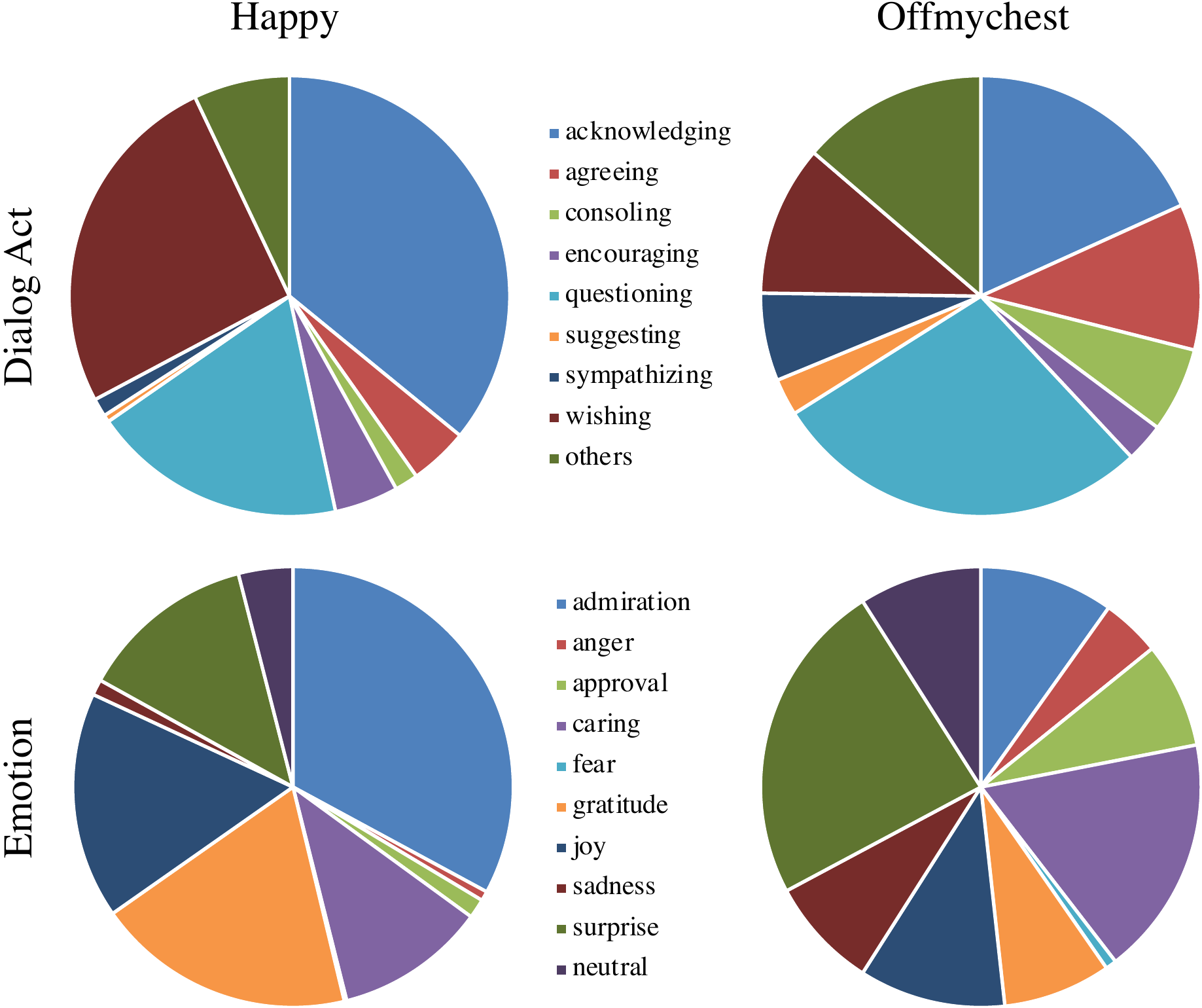}
  \caption{
    Statistics of the annotation results of DA and EM on the two domains.
  }
  \label{fig:stats}
\end{figure}

We computed the proportions of the last responses annotated with ER / IP / EX. 
In the Happy domain, the proportions are 76.0\% / 10.2\% / 18.7\%, while in the Offmychest domain are 57.1\% / 21.4\% / 27.9\% respectively.
The statistics of DA and EM are shown in Figure \ref{fig:stats}.

We can find several differences between two domains.
In terms of \textbf{communication mechanism}, the responses in the Offmychest domain prefer \textit{interpretation} and \textit{exploration}, while \textit{emotional reaction} occupies a larger proportion in the Happy domain.
In terms of \textbf{DA}, the actions that provide support (such as \textit{agreeing}, \textit{consoling}, \textit{suggesting}, and \textit{sympathizing}) are more frequently adopted in the Offmychest domain.
It is similar when it comes to \textbf{emotion}, where the emotions such as \textit{approval} and \textit{caring} are displayed more commonly when responding to the posts with negative sentiments.
We also observed that the responses in the Offmychest domain may also display the emotions like \textit{anger} and \textit{sadness}, indicating that they do understand the experiences and feelings of the conversation partners.

\end{document}